%% file: ltexpprt_doublecolumn.tex
\documentclass[twoside,leqno,twocolumn]{article}

\usepackage[letterpaper]{geometry}

\usepackage{ltexpprt}
\usepackage{hyperref}

\usepackage{newfloat}
\usepackage{listings}
\usepackage{xcolor}
\usepackage{algorithm}
\usepackage{algorithmic}
\usepackage{multicol}
\usepackage{multirow}
\usepackage{amsmath, amssymb, amsfonts}
\usepackage{booktabs}
\usepackage{graphicx} 
\usepackage{bbm}
\usepackage[switch]{lineno}

\begin{document}
%
\newcommand\relatedversion{}

\title{\Large Beyond Models! Explainable Data Valuation and Metric Adaption for Recommendation}
\author{Renqi Jia\thanks{Department of Computer Science, City University of Hong Kong, Hong Kong SAR. Email: renqijia2-c@my.cityu.edu.hk, dawnkun1993@gmail.com, boweihe2-c@my.cityu.edu.hk, \{weitaoxu, chenma\}@cityu.edu.hk} \and 
Xiaokun Zhang\footnotemark[1] \footnotemark[4] \and 
Bowei He\footnotemark[1] \and 
Qiannan Zhu\thanks{School of Artificial Intelligence, Beijing Normal University, China. Email: zhuqiannan@bnu.edu.cn} \and 
Weitao Xu\footnotemark[1] \and 
Jiehao Chen \thanks{China Academy of Industrial Internet, China. Email: chenjiehao@china-aii.com} \and
Chen Ma\footnotemark[1] \thanks{Corresponding author.}}

\date{}

\maketitle


\fancyfoot[R]{\scriptsize{Copyright \textcopyright\ 2025 by SIAM\\
Unauthorized reproduction of this article is prohibited}}






\input{Contents/0_abstract}
\input{Contents/1_introduction}
\input{Contents/2_related_work}
\input{Contents/3_problem_formulation.tex}
\input{Contents/4_preliminary.tex}
\input{Contents/5_method}
\input{Contents/6_experiments}
\input{Contents/7_conclusion}

\bibliography{ltexpprt_doublecolumn}
\end{document}

%% file: Contents/0_abstract.tex
\begin{abstract}
    \small\baselineskip=9pt 
        User behavior records serve as the foundation for recommender systems. While the behavior data exhibits ease of acquisition, it often suffers from varying quality.
        Current methods employ data valuation to discern high-quality data from low-quality data. However, they tend to employ black-box design, lacking transparency and interpretability. Besides, they are typically tailored to specific evaluation metrics, leading to limited generality across various tasks.
        To overcome these issues, we propose an explainable and versatile framework DVR which can enhance the efficiency of data utilization tailored to any requirements of the model architectures and evaluation metrics. For explainable data valuation, a data valuator is presented to evaluate the data quality via calculating its Shapley value from the game-theoretic perspective, ensuring robust mathematical properties and reliability. In order to accommodate various evaluation metrics, including differentiable and non-differentiable ones, a metric adapter is devised based on reinforcement learning, where a metric is treated as the reinforcement reward that guides model optimization.
        Extensive experiments conducted on various benchmarks verify that our framework can improve the performance of current recommendation algorithms on various metrics including ranking accuracy, diversity, and fairness. Specifically, our framework achieves up to 34.7\% improvements over existing methods in terms of representative NDCG metric. The code is available at https://github.com/renqii/DVR.
    \end{abstract}

%% file: Contents/1_introduction.tex
\section{Introduction}
With the vast amount of online information, Internet users are constantly exposed to an ever-growing number of online products or services. This abundance makes it challenging for individuals to find items that truly align with their interests. To solve the problem of information overload, personalized recommendation systems have emerged, playing a pivotal role in helping users navigate through the vast choices in modern society~\cite{ma2019hierarchical, ma2020memory, 10.1145/3477495.3532043, 10.1145/3626772.3657761}.

Current recommender systems usually prioritize the development of intricate model structures like RNN~\cite{HidasiKBT15}, Attention~\cite{10.1145/3626772.3657748}, Transformer~\cite{10.1145/3357384.3357895}, and GNN~\cite{he2020lightgcn}. With delicate model designs, these methods typically resort to Bayesian Personalized Ranking (BPR) to train their models on users' behavior data. BPR generates training samples by negative sampling~\cite{10.5555/1795114.1795167}, constructing triplets $\{\textit{user, positive item, negative item}\}$ to distinguish the positive items from the negative ones. The quality of these training samples plays a crucial role in the BPR training procedure, directly impacting the effectiveness of the recommendation model. However, most methods overlook data quality as they use the vanilla BPR training procedure, which penalizes all samples equally without accounting for distinguishing the distinct effects of various training instances.

A common case of fluctuating data quality arises from the negative sampling process, where the sampled items may include both true negative items and false negatives. For instance, in a movie recommendation system where a user prefers action movies, sampling true negative items like art films leads to higher-quality data over false ones sampled as war movies. Consequently, treating all training samples equally in this context can lead to sub-optimal representation learning, ultimately undermining the model's performance.
To differentiate high-quality data from low-quality data, some data-valuation methods for recommendation have been proposed~\cite{10.1145/3437963.3441800, 10.1145/2556195.2556248,gantner2012personalized,10.1145/3366423.3380187, wu2022adapting}. These methods typically resort to some pre-defined heuristic rules to identify the data quality, such as the prediction scores, popularity levels, or loss values. While these approaches have demonstrated promising performance, we identify two main limitations:
\begin{itemize}
    \item \textbf{Lacking interpretability}. Current methods employ a black-box design for data valuation, making it difficult to understand how data is evaluated and assigned value. This lack of transparency poses challenges in aligning data valuation with its actual contribution to model performance. For instance, there may be cases where data has a minimal impact on model outcomes, yet receives a high valuation. By incorporating explainability into data valuation, it becomes possible to ensure that the valuation accurately reflects data's real impact on model performance. This alignment would enhance both the transparency and reliability of data valuation processes, offering deeper insights into the data value in improving recommendations. 
    \item \textbf{Limited generality}. Prior works concentrate on enhancing recommendations through tailored model designs on certain metrics like accuracy, limiting the scalability of these approaches. There is a growing need for more versatile and adaptable work that can be applied across various model architectures and metrics. Many metrics utilized in performance evaluation are non-differentiable and hold significance in assessing various facets of the recommender system. For instance, Recall is crucial for measuring accuracy, while Category Coverage is instrumental in evaluating diversity. Enhancing the recommendation system across diverse aspects can lead to an increase in user satisfaction. Therefore, it is significant to handle both the differentiable and non-differentiable metrics.
\end{itemize}

To overcome these issues, we propose an explainable and versatile \textbf{\underline{D}}ata \textbf{\underline{V}}aluation framework for \textbf{\underline{R}}ecommendation (DVR) which can enhance the efficiency of data utilization tailored to any requirements of the model architectures and evaluation metrics. For explainable data valuation, a data valuator is presented to evaluate the data quality by measuring the contribution of data to model performance. The data valuator views the data as player and model performance as the outcome to compute the explainable Shapley value from a game-theoretic perspective, which has good mathematical properties and reliability. To improve the computation efficiency of the Shapley value, the data valuator utilizes Harsanyi interaction to reduce the computational complexity from exponential to constant time. With the aim of accommodating various evaluation metrics, a metric adapter is devised based on reinforcement learning to handle the important non-differentiable metrics that are widely used in many aspects of recommendation evaluation. The metric adapter treats the metric as the reinforcement reward to guide the optimization toward optimal metric performance. To ensure simplicity and efficiency, the framework conducts end-to-end data valuation without complex pre-processing or post-processing. 

The main contributions of this work can be summarized as follows:
\begin{itemize}
    \item We present an explainable and versatile framework to enhance data utilization efficiency across various model architectures and evaluation metrics. To our best knowledge, this work marks the first attempt at handling data valuation in terms of interpretability and generality in the recommendation domain.
    \item 
 We propose the explainable data valuation via calculating Shapley value from the game-theoretic perspective with high efficiency, ensuring understanding and trustworthiness of data values. We achieve metric adaption for both differentiable and non-differentiable metrics by reinforcement learning in an end-to-end manner.
    \item 
 Extensive experiments on various benchmarks show that our framework improves the performance of representative recommendation algorithms on various metrics including ranking accuracy, diversity, and fairness. Specifically, our framework achieves up to 34.7\% improvements in terms of the NDCG metric. Further analysis on explainable data valuation demonstrates the transparency and credibility of our framework.
\end{itemize}

%% file: Contents/2_related_work.tex
\section{Related Work}
\subsection{Data valuation}
Originating from game theory, many data valuation methods based on Shapley value have been proposed. Data Shapley \cite{ghorbani2019data} is commonly used for feature attribution tasks, where the prediction performance of all possible subsets is considered to compute the marginal improvement in performance as the data value. However, this method decouples data valuation from predictor model training, limiting their overall performance due to the lack of joint optimization.
Different from the prior works, DVRL \cite{yoon2020data} directly models the data values using learnable neural networks. For training the data value estimator, DVRL utilizes a reinforcement learning approach coupled with a sampling process. DVRL demonstrates model-agnostic behavior and can be applied even to non-differentiable target objectives. Notably, the learning process is conducted jointly for the data value estimator and the associated predictor model, leading to exceptional outcomes across all considered use cases. To efficiently compute the Shapley value of input variables in a deep learning model, HarsanyiNet utilizes the intermediate neurons the network to capture Harsanyi Interaction~\cite{chen2023harsanyinet}, which allows for the precise computation of the Shapley value.

\subsection{Data Valuation for Recommendation}
Aim at making better use of positive pairs, TCE and RCE \cite{10.1145/3437963.3441800} propose a reweighting method to set user-item pairs with larger loss values as noises so that it assigns lower weights to those positive user-item pairs and reduce their training impact. To evaluate the negative samples, the earlier methods use negative sampling to subsample some unobserved items as negatives based on the predefined sampling distributions, such as uniform and popularity distribution~\cite{10.5555/3367243.3367349,caselles2018word2vec, 10.1145/3097983.3098202, gantner2012personalized,zhu2022gain}. To improve the ability to adapt to different model states and users, the adaptive sampling method is proposed to solve this problem by devising additional measures to select hard samples\cite{10.1145/2556195.2556248, 10.1145/3366423.3380187}. Several works also use auxiliary information to guide the negative sampling process, such as knowledge-graph\cite{10.1145/3366423.3380098, 10.5555/3367243.3367349}. Considering the value of both the positive and negative items jointly on a triplet level, TIL \cite{wu2022adapting} formulates the problem of learning data values as a bilevel optimization task, enabling adaptive learning of the data values for training triplets.

%% file: Contents/3_problem_formulation.tex
\section{Problem Formulation}
In this paper, the recommendation task takes the user behavior data as input. 
Let $\mathcal{U}$ and $\mathcal{V}$ be sets of users and items respectively, where $|\mathcal{U}|= m$, and $|\mathcal{V}|= n$. We use the index $u \in \mathcal{U}$ to denote a user and $i\in \mathcal{V}$ to denote an item. 
The user-item rating matrix is denoted as $\mathbf{R} = [r_i^u]^{m\times n}\in \mathbb{R}^{m\times n}$ to indicate whether user $u$ has interacted with item $i$, where $r_i^u=1$ represents user $u$ has interacted with item $i$,  whereas $r_i^u=0$ represents user $u$ has not interacted with item $i$. 
We use $\mathcal{V}^{+}_{u}=\{i\in\mathcal{V}|r_i^u=1\}$ to represent a set of items that user $u$ has interacted with.
$\mathcal{V}^{+}_{u}$ can be splited into a training set $\mathcal{S}_{u}^{+}$ and a testing set $\mathcal{T}_{u}$, requiring that $\mathcal{S}_{u}^{+} \cup \mathcal{T}_{u} = \mathcal{V}^{+}_{u}$ and $\mathcal{S}_{u}^{+} \cap \mathcal{T}_{u} = \emptyset$. It worth noted that $\mathcal{S}_u^{-}=\{i|r_i^u=0,i\in \mathcal{I}\}$, which means $\mathcal{S}_u^{-}$ consists of the negative items that user $u$ have not interacted with. The training set is denoted as $\mathcal{D}=\{(u,i, j)|u\in \mathcal{U}, i \in \mathcal{S}_{u}^{+}, j \in \mathcal{S}_{u}^{-}\}$. The testing set is denoted as $\mathcal{\hat{D}}=\{(u,i)|u\in \mathcal{U}, i \in \mathcal{T}_{u}\}$.

In the recommendation task, the model aims to recommend a list of $k$ items $\mathcal{X}_u$ for the user $u$, which matches the condition $\mathcal{X}_u\cap \mathcal{S}_u^{+}=\emptyset$. 
By comparing the recommendation list $\mathcal{X}_u$ with the testing set $\mathcal{T}_u$, we evaluate the recommendation quality from various perspectives, including accuracy, diversity, and fairness. 

%% file: Contents/4_preliminary.tex
\section{Preliminary}
For optimization of the recommendation model, the Bayesian Personalized Ranking (BPR) is used to learn the user preference from behavior data. The central idea of BPR is to maximize the ranking of positive items compared with the randomly sampled negative ones, achieved by the following loss function: 
\begin{equation}
    \mathcal{L}_\text{BPR}(u,i,j;\Theta)=-log\sigma(f(u,i;\Theta) - f(u,j;\Theta)),
\end{equation}
where $(u,i,j)$ is the training sample with a positive item $i$ and a negative item $j$ for user $u$. $f(\Theta)$ refers to the recommendation model. The learnable parameter $\Theta$ includes the user embedding $\mathbf{p}_u \in \mathbb{R}^d$ and the item embedding $\mathbf{q}_i \in \mathbb{R}^d$, where $d$ is the embedding dimension. $f(u,i;\Theta)$ is used to compute the relevance score between user $u$ and item $i$.

%% file: Contents/5_method.tex
\begin{figure*}[h]
    \centering
    \includegraphics[scale=0.6]{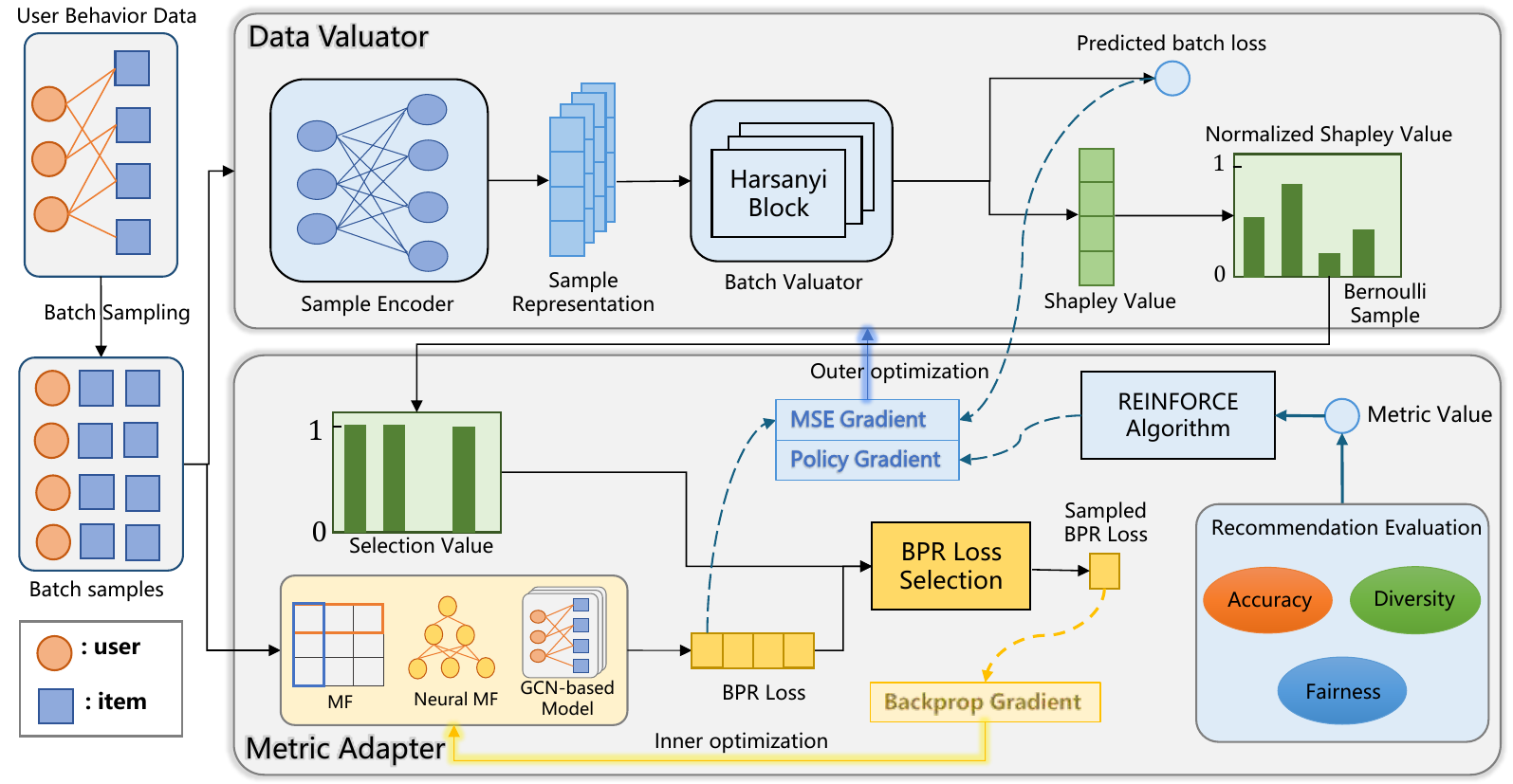}
    \caption{The architecture of the whole DVR framework. The left side of the figure is batch samples from behavior data as the input of the framework. The framework consists of the inner optimization for the recommendation model and the outer optimization of the data valuator. The yellow and blue glowing two-bend arrows mean the backward propagation to optimize the recommendation model and reinforcement learning to optimize the data valuator, respectively.}
    \label{fig:framework}
\end{figure*}

\section{Methodology}
In this section, we first introduce the data valuator for explainable data valuation that evaluates the data quality by calculating the Shapley value from the game-theoretic perspective. Then, we present the metric adapter that achieves metric adaption for both differentiable and non-differentiable metrics by reinforcement learning in an end-to-end manner.

\subsection{Data Valuator}

Our intuition for explainable data valuation is that a sample's data quality is related to its impact on model performance. That is, samples that contribute more to model performance are considered high-quality data, whereas those with less impact are deemed low-quality data. Therefore, the sample's contribution to the model performance can be viewed as the explanation of the data valuation. 

To incorporate the above intuition, we propose to measure the contribution of each training sample $(u, i, j)$ to the model performance as the data value $w_{uij}$. As samples work together to impact the model's performance collectively, we contend that considering the training sample collectively is better than measuring them in isolation. Based on game theory, the Shapley value~\cite{winter2002shapley} is a popular technique that computes the players' contribution to the outcome by the average marginal contribution considering all possible coalitions. Therefore, we treat each sample as the player cooperating with others and contributing to model performance.

Given the large volume of training samples in the recommender system, the computational complexity associated with computing all potential sample coalitions with the original Shapley value is prohibitively high. To reduce computational complexity, we focus on computing the Shapley value of samples within the training batch data, which is represented as $\mathcal{B}= \{(u_1, i_1, j_1), (u_2, i_2, j_2), ..., (u_{n_b}, i_{n_b}, j_{n_b})\}$. The Shapley value for sample $h_m =(u_m, i_m, j_m)\in \mathcal{B}$ is formulated as:
\begin{equation}
\small
\begin{aligned}
\phi\left(h_m\right)=\sum_{\mathcal{S}\subseteq \mathcal{B}\setminus \{h_m\}}\frac{\left|\mathcal{S}\right|!\left(n_b-\left|\mathcal{S}\right|-1\right)!}{n_b!}\left[v\left(\mathcal{S}\cup \{h_m\}\right)-v\left(\mathcal{S}\right)\right],
\end{aligned}
\end{equation}
where $v(\mathcal{S})$ is the value function that computes the model performance with the sample set $\mathcal{S}$ as the input. Many data valuation methods~\cite{ghorbani2019data} utilize retraining techniques to calculate $v(\mathcal{S})$. However, due to the large computational consumption, it is challenging to apply the retraining-based methods to the scenario of the recommender system.

For the scenario of the recommender system, we employ a customized DNN consisting of Harsanyi Interaction to compute the Shapley value effectively. Specifically, Harsanyi interaction $\mathcal{S}$ represents the sample coalition contributing to the model performance collectively. Each Harsanyi interaction $\mathcal{S}$ makes a specific numerical contribution, denoted by $I(\mathcal{S})$, to the model performance. Based on prior research~\cite{harsanyi1982simplified}, the Shapley value is proved to be calculated as:
\begin{equation}
    \label{eq:interaction}
    \begin{aligned}
        \phi(h_m)=\sum_{\mathcal{S}\subseteq \mathcal{B}:\mathcal{S}\ni h_m}\frac1{|\mathcal{S}|}I(\mathcal{S}).
    \end{aligned}
\end{equation} The customized DNN is specified as $w_{\mathcal{B}}(\Lambda)=\mathcal{G}(\mathcal{B};\Lambda)$, where $\Lambda$ is a set of learnable parameters.

\subsubsection{Sample Encoder}
We first employ the sample encoder to encode each sample within the training batch to obtain the sample representation. For the training sample $h_m\in \mathcal{B}$, we use the following network get the representation of $h_m$:
\begin{equation}
    \begin{aligned}
        \hat{\mathbf{e}}_{m} &= \text{ReLU}(W_1\cdot  \mathbf{e}_{m}+b_1),\\
        \mathbf{h}_{m} &= \text{Tanh}(W_2 \cdot \hat{\mathbf{e}}_{m}+b_2),
    \end{aligned}
\end{equation}
where the input $\mathbf{e}_{m}=[\mathbf{p}_{u_m}; \mathbf{q}_{i_m}; \mathbf{q}_{j_m}]\in \mathbb{R}^{3d}$ is the concatenation of the representation of user $u_m$, item $i_m$, and item $j_m$. $W_1\in \mathbb{R}^{d \times 3d}$, $W_2\in \mathbb{R}^{d}$, $b_1\in \mathbb{R}^{d}$, and $b_2\in \mathbb{R}$. We take $\mathbf{h}_{m}$ as the sample representation.

\subsubsection{Batch Valuator}
To compute the contribution of each training sample to the model performance, the batch valuator is achieved by a customized DNN, which takes batch samples as input to predict the model performance. 

Referring to HarsanyiNet~\cite{chen2023harsanyinet}, the customized DNN is designed by considering the AND relationship between the children nodes of the neuron. Since the customized DNN contains $L$ stacked Harsanyi blocks, we focus on the $m$-th neuron in the $l$-th block. Given the children set $\mathcal{C}_m^{(l)}$, the neural activation $\mathbf{z}_m^{(l)}$ of the neuron $(l, m)$ is computed by applying the AND operation on the Linear network:
\begin{equation}
    \begin{aligned}
        \hat{\mathbf{z}}_m^{(l)}&=\mathbf{A}_m^{(l)}\cdot (\textstyle{\sum_m^{(l)}}\cdot \mathbf{z}^{(l-1)}),\\
        \mathbf{z}_m^{(l)}&=\text{ReLU}(\hat{\mathbf{z}}_m^{(l)}\cdot\prod_{(l,m)\in\mathcal{C}_{m}^{(l)}}\mathbbm{1}(\mathbf{z}_{m}^{(l)}\neq 0)).
    \end{aligned}
\end{equation}
 The input of the $l$-th block are neurons from the last block, denoted as $\mathbf{z}^{(l-1)}=[\mathbf{z}_1^{(l-1)},...,\mathbf{z}_m^{(l-1)}...,\mathbf{z}_{M^{(l-1)}}^{(l-1)}]$ where $M^{(l-1)}$ is the number of neurons in the $(l-1)$-th block. Especially, for the first block, the input neurons are $\mathbf{z}^{(0)}=[\mathbf{h}_1, ...,\mathbf{h}_m,... \mathbf{h}_{n_b}]$. The children set $C_m^{(l)}$ is implemented as a trainable binary diagonal matrix $\textstyle{\sum_m^{(l)}}\in \{0,1\}^{M^{(l-1)}\times M^{(l-1)}}$, which selects children nodes of the neuron $(l, m)$ from the neurons in the last blocks. $\mathbf{A}_m^{(l)}\in \mathbb{R}^{M^{(l-1)}}$ denotes the weight vector. 
 
\subsubsection{Prediction and Optimization}
Neurons in multiple layers of blocks are used to predict the model performance:
\begin{equation}
    \begin{aligned}
        \hat{y} = \sum_{l=1}^{L}(\mathbf{v}^{(l)})^{T}\mathbf{z}^{(l)},
    \end{aligned}
\end{equation}
where $\mathbf{v}^{(l)}=[\mathbf{v}^{(l)}_1, ..., \mathbf{v}^{(l)}_{M^{(l)}}]\in \mathbb{R}^{M^{(l)}}$ denotes the weight vector. $\mathbf{z}^{(l)}=[\mathbf{z}^{(l)}_1, ..., \mathbf{z}^{(l)}_{M^{(l)}}]$ denotes neurons in the $l$-th block. $\hat{y}$ is the prediction value of model performance. We use batch loss to represent the model performance. Then, the model is trained by the mean squared error of predicted value $\hat{y}$ and the batch BPR loss. The loss function is denoted as:
\begin{equation}
\label{eq:mse}
    \begin{aligned}
        \mathcal{L}_\text{MSE}(\mathcal{B};\Lambda) = (\hat{y} - \sum_{(u,i,j) \subset \  \mathcal{B}}\mathcal{L}_{\text{BPR}}(u,i,j;\Theta))^2.
    \end{aligned}
\end{equation}

\subsubsection{Calculation of Shapley Value}
As proved in the prior method~\cite{chen2023harsanyinet}, through the Harsanyi Interaction represented by neurons in the customized DNN, we implement the formula \ref{eq:interaction} to compute Shapley value $\phi(h_m)$ as:
\begin{equation}
    \label{eq:svcal}
    \begin{aligned}
        \mathcal{\hat{C}}_m^{(l)} &:= \cup_{(l,m)\in \mathcal{C}_m^{(l)}}\mathcal{\hat{C}}_{m}^{(l)}, \quad s.t. \mathcal{\hat{C}}_{m}^{(1)}:=\mathcal{C}_m^{(1)},\\
    \phi(h_m)&=\sum_{l=1}^L\sum_{m=1}^{M^{(l)}}\frac1{|\mathcal{\hat{C}}_m^{(l)}|}\mathbf{v}_{m}^{(l)}\mathbf{z}_{m}^{(l)}\mathbbm{1}(\mathcal{\hat{C}}_m^{(l)}\ni h_m).
    \end{aligned}
\end{equation}

\subsection{Metric Adapter}
\subsubsection{The Bilevel Optimization}
We utilize the assigned Shapley value for training sample selection. For the training batch $\mathcal{B}$, the probability of sampling is denoted as $\pi(\mathcal{B},\mathbf{s};{\Lambda}) = \prod_{i=1}^{|\mathcal{B}|}[\hat{w}_{uij}({\Lambda})^{s_{uij}} \cdot (1-\hat{w}_{uij}({\Lambda}))^{1-s_{uij}}]$. In order to calculate the sampling probability, we normalize the Shapley value as $\hat{w}_{uij}$. The selection vector is dentoed as $\mathbf{s}=[s_{uij}]^{|\mathcal{B}|}$ where $s_{uij}$ is the parameterized Bernoulli variable with the probability $\hat{w}_{uij}$ to be 1 and $1-\hat{w}_{uij}$ to be 0. If $s_{uij}=1/0$, the sample  $(u,i,j)$ is selected/not selected for training the recommendation model.

After data selection, we use the selected data to train the recommendation model and data valuator in a bilevel manner. The bi-level optimization is formulated as:
\begin{equation}
    \begin{aligned}
    &\min_{\Lambda}\mathbb{E}_{s \sim \pi(\mathcal{B},\cdot;{\Lambda})} \mathcal{R}(\hat{\mathcal{D}}, f(\Theta^*(\Lambda))) + \mathcal{L}_\text{MSE}(\mathcal{B};\Lambda) \\
    &\text{s.t.}\Theta^{*}(\Lambda)=\mathop{\arg\min}_{\Theta}\mathbb{E}_{s \sim \pi(\mathcal{B},\cdot;{\Lambda})} \mathcal{L}_{\text{BPR}}(u,i,j;\Theta).       
    \end{aligned}
    \label{eq:bilevel2}
\end{equation}
The recommendation evaluator $\mathcal{R}(\hat{\mathcal{D}}, f(\Theta^*(\Lambda)))$ means that $\mathcal{R}(\cdot)$ takes the test data $\hat{\mathcal{D}}$ and recommendation model $f(\Theta^*(\Lambda))$ as input to measure the metric $R$. In the outer optimization, the data valuator $\mathcal{G}(\mathcal{B};\Lambda)$ is optimized by the performance metric and the MSE loss. In the inner optimization, the recommendation model is optimized by the BPR loss of selected training samples.

\subsubsection{Reinforced Metric Adaption}
The sampled BPR loss and MSE loss can be optimized with backprop gradients~\cite{amari1993backpropagation} because they are differentiable. The challenge arises from the fact that many metrics are non-differentiable, which poses difficulties for optimization of the metric in the outer optimization of Eq. (\ref{eq:bilevel2}):
\begin{equation}
    \begin{aligned}
        \hat{l}(\Lambda)=\mathbb{E}_{s \sim \pi(\mathcal{B},\cdot;{\Lambda})} \mathcal{R}(\hat{\mathcal{D}}, f(\Theta^*(\Lambda))).
    \end{aligned}
    \end{equation}

To handle this situation, we propose to use REINFORCE algorithm to optimize non-differentiable metrics. The optimization gradient $\nabla_{\Lambda}\hat{l}(\Lambda)$ can be computed directly as:
\begin{equation}
    \label{eq:metricpg}
    \small
    \begin{aligned}
        &\nabla_{\Lambda}\hat{l}(\Lambda)
        &=\mathbb{E}_{s\sim \pi(\mathcal{B},\cdot;{\Lambda})}[\mathcal{R}(\hat{\mathcal{D}}, f(\Theta^*(\Lambda)))\cdot \nabla_{\Lambda}log(\pi(\mathcal{B},s;{\Lambda}))],
    \end{aligned}
\end{equation}
where $\mathcal{R}(\hat{\mathcal{D}}, f(\Theta^*(\Lambda)))\nabla_{\Lambda}log(\pi(\mathcal{B},s;{\Lambda}))$ is the policy gradient of $ \nabla_{\Lambda}\hat{l}(\Lambda)$, which can gudide the data valuator to identify samples that are beneficial for the metric. In the Eq. \ref{eq:metricpg}, $\nabla_{\Lambda}log(\pi(\mathcal{B},s;{\Lambda}))$ can be further computed as:
\begin{equation}
    \small
    \begin{aligned}
    \label{eq:lambda}
        &\nabla_{\Lambda}log(\pi(\mathcal{B},s;{\Lambda}))
        \\&=\nabla_{\Lambda}\sum_{(u,i,j)\in \mathcal{B}} log(\hat{w}_{uij}(\Lambda)^{s_{uij}}\cdot(1-\hat{w}_{uij}(\Lambda))^{1-s_{uij}})
        \\&=\sum_{(u,i,j)\in \mathcal{B}} s_{uij}\nabla_{\Lambda}log(\hat{w}_{uij}(\Lambda))+(1-s_{uij})\nabla_{\Lambda}log(1-\hat{w}_{uij}(\Lambda)).
    \end{aligned}
\end{equation}
Therefore, given the optimum $\Theta^{*}(\Lambda)$ of the recommendation model, we can update the parameter $\Lambda$ of the data valuator as:
\begin{equation}
    \begin{aligned}
    \label{eq:pge}
        \Lambda \longleftarrow \Lambda - \eta [\mathcal{R}(\hat{\mathcal{D}}, f(\Theta^*(\Lambda)))\nabla_{\Lambda}log(\pi(\mathcal{B},s;{\Lambda})].
    \end{aligned}
\end{equation}
It is clear that Eq. (\ref{eq:pge}) does not involve any implicit differentiation because its component $\mathcal{R}(\hat{D}, f(\Theta^*(\Lambda)))$ and $log(\pi(\mathcal{B},s;{\Lambda})$ can be computed via forward propagation and model evaluation without the need for backpropagation. Therefore, we can update $\Lambda$ via Eq. (\ref{eq:pge}) very efficiently. 
Hence, we can solve our bilevel optimization problem by iteratively optimizing the objective functions of both the recommender and the data valuator. 

%% file: Contents/6_experiments.tex
\section{Experiment}
We compare our framework on four public datasets with various base models to sate-of-the-art baselines from the aspects of accuracy, diversity, and fairness.
\subsection{Datasets}
We use four real-world datasets in our experiments. Table \ref{tab:DataStatistics} provides an overview of the data statistics. \textbf{\textit{Beauty}} and \textbf{\textit{CD}} are both product recommendation datasets adopted from the \textit{Amazon Review Dataset}\cite{he2016ups}. It covers users' purchases over the category of \textit{All Beauty} and \textit{CDs and Vinyl} with rating score. \textbf{\textit{LastFM}} was collected from the \textit{Last.fm} music social platform \cite{Celma:Springer2010}, covering a large amount of users' music listening activities and metadata from the period between 2005 and 2009. \textbf{\textit{Gowalla}} was collected worldwide from the \textit{Gowalla} website \cite{cho2011friendship}, which is a location-based social networking website over the period from February 2009 to October 2010.

When dealing with datasets originally containing explicit ratings, we consider any ratings equal to or greater than four (on a scale of five) as positive feedback and all other ratings as missing entries to maintain consistency within the implicit feedback setting. Furthermore, to ensure data quality and reliability, we follow the common practice used in prior works to filter out users and items with fewer than ten ratings. 

\begin{table}[h]
    \centering
    \small
    \caption{The statistics of four read-world datasets.}
    \label{tab:DataStatistics}
    \begin{tabular}{lllll}
        \toprule
                & \textit{Beauty}  & \textit{CD} & \textit{LastFM}  & \textit{Gowalla} \\ \midrule
       
        \#User    & 8,159    & 11,346    & 23,385   & 29,858   \\ \hline
        \#Item    & 5,862    & 32,705    & 34,186   & 40,981   \\ \hline
        \#Interaction & 98,566   & 466,501  & 982,798  & 1,027,370 \\ \hline
        Density   & 0.206\% & 0.126\%  & 0.123\% & 0.084\% \\ \bottomrule
    \end{tabular}
\end{table}

\subsection{Evaluation Protocols}
We utilize cross-validation to evaluate our proposed model. The user-item interactions are divided into three sets: training set, validation set, and testing set, with a ratio of 8:1:1, respectively. We evaluate our model from three aspects: accuracy, diversity, and fairness. With respect to accuracy, Recall and Normalized Discounted Cumulative Gain (NDCG) are used. Recall@K measures how many target items are retrieved in the recommendation result, while NDCG@K further takes into account their positions in the ranking list \cite{jarvelin2002cumulated}. To evaluate the diversity, we use Category Coverage (CC) and Intra-List Distance (ILD). CC@k computes the proportion of unique categories represented in the top-K recommendation results. ILD@k calculates the average dissimilarity between item pairs within the recommended item list. For fairness, we use the Gini index to calculate the imbalance in the item distribution within the recommendation results. It is worth noting that we consider the ranking list of all items, excluding the training items in the user history, rather than ranking a smaller set of random items alongside the target items, as suggested in recent research \cite{li2020sampling}. All experiments are run five times with the same seed to control the data partition.

\begin{table*}[h]
    \tabcolsep 0.06in
    \centering
    \small
    \caption{The performance comparison of all methods on the backbone of BPR in terms of R@20 (Recall@20) and N@20 (NDCG@20) in percentage (\%).}
    \label{tab:OverallResult}
    \begin{tabular}{c|c|c|c@{\extracolsep{4pt}}c@{\extracolsep{4pt}}c@{\extracolsep{4pt}}c@{\extracolsep{4pt}}c@{\extracolsep{4pt}}c@{\extracolsep{4pt}}c@{\extracolsep{4pt}}c|c@{\extracolsep{4pt}}c@{\extracolsep{4pt}}c@{\extracolsep{4pt}}}
    \hline
    \textbf{Datasets}                  & \textbf{backbone}                  & \textbf{Metric} & \textbf{BPR}   &  \begin{tabular}[c]{@{}l@{}}\textbf{AO}\\ \textbf{-BPR}\end{tabular} &  \begin{tabular}[c]{@{}l@{}}\textbf{W}\\ \textbf{-BPR}\end{tabular} & \textbf{PRIS} & \begin{tabular}[c]{@{}l@{}}\textbf{TCE}\\ \textbf{-BPR}\end{tabular} & \begin{tabular}[c]{@{}l@{}}\textbf{RCE}\\ \textbf{-BPR}\end{tabular} & \begin{tabular}[c]{@{}l@{}}\textbf{TIL}\\ \textbf{-UI}\end{tabular} & \begin{tabular}[c]{@{}l@{}}\textbf{TIL}\\ \textbf{-MI}\end{tabular} & \begin{tabular}[c]{@{}l@{}}\textbf{DVR}\\ \textbf{-Loss}\end{tabular} & \begin{tabular}[c]{@{}l@{}}\textbf{DVR}\\ \textbf{-Recall}\end{tabular} & \begin{tabular}[c]{@{}l@{}}\textbf{DVR}\\ \textbf{-NDCG}\end{tabular}  \\ \hline
    \multirow{8}{*}{Beauty} & \multirow{2}{*}{\textbf{BPRMF}}       & R@20   & 11.33  & 11.36  & 11.35 & 11.64 & 12.20 & 12.84 & 12.96 & \underline{13.47} & 15.82& 15.83& \textbf{15.84}\\
                              &                           & N@20   & 5.47  & 5.51  & 5.49 & 5.65 & 6.41 & 6.98 & 7.32  & \underline{7.51} & \textbf{8.76}& 8.67& 8.70\\
                              & \multirow{2}{*}{\textbf{NeuMF}}    & R@20   & 10.23  & 10.25 & 10.23 & 11.17 & 11.25 & 12.18 & 12.59 & \underline{12.81} & 13.41& 13.47& \textbf{13.50}\\
                              &                           & N@20   & 5.08  & 5.10 & 5.09 & 5.92 & 6.01 & 6.23 & 6.54 & \underline{6.75} & 7.19& 7.40& \textbf{7.48}\\ \cline{2-14} 
                              & \multirow{2}{*}{\textbf{MGCF}}     & R@20   & 11.41 & 11.41 & 11.42 & 11.78 & 12.38 & 12.92 & 13.42 & \underline{14.20} &  15.65&  15.67&   \textbf{15.68}\\
                              &                           & N@20   &  5.59 & 5.63 & 5.64 & 5.72 & 6.60 & 6.88 &  7.51 & \underline{7.83} &   \textbf{8.70}&  8.66&  8.67\\
                              & \multirow{2}{*}{\textbf{LightGCN}} & R@20   & 11.21 & 11.24 & 11.25 & 11.55 & 12.32 & 12.84 & 13.47 & \underline{14.46} &  16.40&  \textbf{16.43}&   16.42\\
                              &                           & N@20   & 5.33  & 5.35 & 5.36 & 5.69 & 6.54 & 6.69 & 7.54 & \underline{8.05}  &   8.94&  8.96&  \textbf{8.97}\\ \hline
    \multirow{8}{*}{CD}   & \multirow{2}{*}{\textbf{BPRMF}}       & R@20   & 9.99 & 10.03  & 10.01 & 10.32 & 9.03 & 9.71 & 10.88   & \underline{12.03}   & 15.02& \textbf{15.16}& 15.08\\
                              &                           & N@20   & 5.81  & 5.83  & 5.92 & 6.04  & 5.36 & 5.73 & 6.30 & \underline{7.11}  & 8.98& \textbf{8.99}& 8.97\\
                              & \multirow{2}{*}{\textbf{NeuMF}}    & R@20   & 11.03 & 11.08  & 11.05 & 11.93 & 11.77 & 11.42 & 11.85   & \underline{13.33}   & 14.10& \textbf{14.12}   & 14.10\\
                              &                           & N@20   & 6.40  & 6.43  & 6.62 & 7.22 & 6.86 & 6.77 & 7.01   & \underline{7.83}   & 8.48& 8.50& \textbf{8.52}\\ \cline{2-14} 
                              & \multirow{2}{*}{\textbf{MGCF}}     & R@20   & 13.80  & 13.86  & 14.01 & 14.24 & 13.82 & 13.94 & 14.30  & \underline{14.67}  &  \textbf{15.21}&  15.20&  15.15\\
                              &                           & N@20   & 8.02  & 8.08  & 8.07 & 8.24  & 8.20 & 8.11 & 8.43   & \underline{8.68}   &  9.20&  9.18&  \textbf{9.21}\\
                              & \multirow{2}{*}{\textbf{LightGCN}} & R@20   & 13.40 & 13.44  & 13.52 & 13.77 & 13.52 & 13.45 & 13.92  & \underline{14.21}   &  \textbf{14.98}&  14.95&  14.96\\
                              &                           & N@20   & 7.90  &  7.95  & 7.96  & 8.15  & 8.12 & 7.98 & 8.34    & \underline{8.61}   &  9.27&  9.26&  \textbf{9.28}\\ \hline
    
    \multirow{8}{*}{LastFM}   & \multirow{2}{*}{\textbf{BPRMF}}       & R@20   & 17.35 & 17.38  & 18.99 & 20.28 & 20.30& 18.50& 20.70   & \underline{21.63}   & 25.86& \textbf{25.96}& 25.92\\
                              &                           & N@20   & 10.47 & 10.52  & 11.43 & 12.07 & 11.91 & 11.46 & 12.48   & \underline{13.22}  & 16.44& \textbf{16.56}& 16.44\\
                              & \multirow{2}{*}{\textbf{NeuMF}}    & R@20   & 19.17 & 19.18  & 20.80 & 21.15 & 21.50 &  21.46 & 22.86   & \underline{23.15}   & 24.91& 25.22& \textbf{25.26}\\
                              &                           & N@20   & 11.63 & 11.67 & 12.15 & 12.26 & 13.07 & 12.99 & 14.10  & \underline{14.21}  & 15.80& \textbf{15.99}& 15.96\\ \cline{2-14} 
                              & \multirow{2}{*}{\textbf{MGCF}}     & R@20   & 21.26 & 21.32  & 21.28 & 21.97 & 21.80 & 21.22 & 23.14   & \underline{23.44}  &  25.53&  \textbf{25.67}&  25.61\\
                              &                       & N@20   & 13.06 & 13.33  & 13.30 & 13.50 & 13.47 & 13.19 & 14.19  & \underline{14.35}   &   16.12&  16.22&     \textbf{16.25}\\
                              & \multirow{2}{*}{\textbf{LightGCN}} & R@20    & 23.31 & 23.33 & 23.35 & 23.92 & 23.82 & 23.30 & 24.14  & \underline{24.85}   &  27.55&  \textbf{27.61}&  27.60\\
                              &                           & N@20   & 14.25 & 14.28  & 14.26 & 14.68 & 14.51 & 14.20 & 14.82   & \underline{15.35}   &  18.03&  18.09&  \textbf{18.10}\\ \hline
    \multirow{8}{*}{Gowalla}  & \multirow{2}{*}{\textbf{BPRMF}}       & R@20   & 12.19 & 12.22  & 13.02  &  14.88 & 13.94 & 13.82 & 15.80  & \underline{16.42}   & \textbf{17.85}& 17.15& 17.16\\
                              &                           & N@20   & 7.76  & 7.80  & 7.92 & 9.49 & 8.53 & 8.25 & 9.77   & \underline{10.09}   & \textbf{13.59}& 13.02& 13.02\\
                              & \multirow{2}{*}{\textbf{NeuMF}}    & R@20   & 13.55 & 13.59  & 14.35 & 15.83 & 14.80 & 14.76 & {16.89}   & \underline{17.32}   & 16.40   & \textbf{16.42}      & 16.41    \\
                              &                           & N@20   & 8.47  & 8.52  & 8.60 & 9.23 & 8.96 & 8.87 & 10.90   & \underline{11.17}   & 12.46   & 12.47      & \textbf{12.48}    \\ \cline{2-14} 
                              & \multirow{2}{*}{\textbf{MGCF}}     & R@20   & 15.75 & 15.84 & 15.83 & 16.35 & 15.93 & 15.80 & 17.33 & \underline{18.25} &  18.47&  18.52&   \textbf{18.53}\\
                              &                           & N@20   & 9.70  & 9.81  & 9.82 & 9.94 & 10.00 & 9.79 & 10.24  & \underline{11.06} &  14.36&    14.38&   \textbf{14.40}\\
                              & \multirow{2}{*}{\textbf{LightGCN}} & R@20   & 17.73 & 17.75  & 17.78 & 18.03 & 17.99 & 17.82 & 18.01   & \underline{18.61}   &  18.62&  18.85&  \textbf{18.87}\\
                              &                           & N@20   & 11.16 & 11.21  & 11.18 & 11.30 & 11.21 & 11.15 & 11.27   & \underline{11.62}    &  14.47&  14.55&  \textbf{14.58}\\ \hline
    \end{tabular}
\end{table*}

\subsection{Methods Studied}
To show the compatibility of our method, we apply the DVR framework on four recommendation backbones, i.e., BRPMF~\cite{koren2009matrix}, NeuMF~\cite{he2017neural}, MGCF~\cite{8970709}, and LightGCN~\cite{he2020lightgcn}.
Based on these backbones, our DVR framework can optimize diverse metrics, where ranking accuracy is denoted as DVR-Loss, DVR-Recall, and DVR-NDCG, diversity and fairness are labeled as DVR-CC, DVR-ILD, and DVR-Gini. Besides, we compare our framework with various data valuation methods for recommendations including BPR~\cite{10.5555/1795114.1795167}, AOBPR \cite{10.1145/2556195.2556248}, WBPR \cite{gantner2012personalized}, PRIS \cite{10.1145/3366423.3380187}, TCE-BPR, RCE-BPR\cite{10.1145/3437963.3441800}, TIL-UI and TIL-MI \cite{wu2022adapting}.

\subsection{Performance Comparison}
\subsubsection{Comparison of Accuracy}
As shown in Table \ref{tab:OverallResult}, we compare the accuracy performance of DVR-Loss, DVR-Recall, and DVR-NDCG with several baseline approaches. The following is observations about the results: (i) Our framework outperforms most baseline methods across diverse evaluation metrics and consistently maintains this superiority across multiple datasets. This shoes the ability of our data valuator to accurately assess the quality of user behavior data, enhancing the efficacy of the recommendation model.
(ii) Compared to TCE-BPR and RCE-BPR, our framework consistently outperforms them. These approaches assume that noise only comes from positive samples and treat all negative items equally, which is suboptimal. Moreover, relying solely on loss values for weighting positive samples can reinforce errors.
(iii) Our framework outperforms AOBPR, WBPR, and PRIS. One factor is that these methods rely solely on negative samples to assess data quality. They select or adjust the weight of informative negative samples based on manual rules like item popularity or training loss, which may not be comprehensive.
(iv) In comparison to TIL-UI and TIL-MI, our proposed framework can achieve highly competitive performance when applied to both the simple base model (e.g., MF) and complex GNN-based model (e.g., MGCF and LightGCN). This strongly suggests the effectiveness of directly optimizing the accuracy metrics, i.e. Recall and NDCG.
(v) DVR-Loss, DVR-Recall, and DVR-NDCG exhibit similar outcomes, likely because these metrics all focus on accuracy. DVR-Recall and DVR-NDCG generally outperform DVR-Loss, as Recall and NDCG take into account detailed ranking information, while the latter only considers the relationship between the used positive and negative samples.

\subsubsection{Comparison of Diversity}
We compared DVR-CC and DVR-ILD with the BPR model on the Beauty dataset based on BPRMF and MGCF backbones. The performance results are presented in Table \ref{tab:Diversity}. Key observations from the experiments include: (i) DVR-CC and DVR-ILD achieved the best performance for CC@20 and ILD@20 metrics, respectively, as they are directly optimized for diversity metrics. This allows the data valuator to adjust data values for improved training of the recommendation model towards higher diversity. (ii) An interesting finding is that the Recall@20 and NDCG@20 of DVR-CC and DVR-ILD show no significant decrease compared to the BPR model, demonstrating the robustness of our reinforcement learning framework. This is attributed to our approach of enhancing recommendation model optimization by adjusting data values instead of directly incorporating diversity regularizers, which could potentially compromise recommendation performance. (iv) The MGCF outperforms the BPRMF as the backbone for BPR, DVR-CC, and DVR-ILD. This indicates that GCN-based approaches serve as strong backbones suitable for the joint optimization of data valuation and recommendation models.

\begin{table}[t]
\centering
\small
\caption{The performance comparison of BPR and our proposed methods on the Beauty dataset in terms of Recall@20, NDCG@20, CC@20, and ILD@20 in percentage (\%).}
\label{tab:Diversity}
    \begin{tabular}{l|l|l|l|l}
    \hline
    backbone                  & Metric & BPR & DVR-CC & DVR-ILD \\ \hline
    \multirow{2}{*}{BPRMF}       & R@20  & \textbf{11.33}&  11.01&  10.79\\
                               & N@20 & \textbf{5.47}&   5.18&  5.05\\
                               & CC@20 &  69.75&   \textbf{71.11}&   70.04\\
                               & ILD@20 &  90.31&  91.43&  \textbf{91.88}\\ \cline{1-5} 
    \multirow{2}{*}{MGCF} & R@20  & \textbf{11.41}&  11.26&  11.17\\
                               & N@20 & \textbf{5.59}&  5.45&  5.41\\
                               & CC@20  & 71.32&   \textbf{72.85}&  72.83\\
                              & ILD@20 & 91.49&   92.67&  \textbf{92.96}\\ \hline
    \end{tabular}
\end{table}
    
\begin{table}[t]
\small
\centering
\caption{The performance comparison of BPR and our proposed methods in terms of Recall@20, NDCG@20, Gini@20 (Gini Index) in percentage (\%).}
\label{tab:ItemFairness}
\begin{tabular}{l|l|l|l|l}
\hline
Datasets                  & backbone                  & Metric & BPR & DVR-Gini \\ \hline
\multirow{4}{*}{LastFM} & \multirow{2}{*}{BPRMF}       & R@20  &  \textbf{17.35}&  17.21\\
                            &                           & N@20 &  \textbf{10.46}&  10.38\\
                            &                           & Gini@20  &  97.94&  \textbf{95.47}\\ \cline{2-5} 
                            & \multirow{2}{*}{LightGCN} & R@20  &  \textbf{23.31}&  22.37\\
                            &                           & N@20 &  \textbf{14.25}&  13.69\\
                            &                           & Gini@20  &  97.86&  \textbf{96.23}\\ \hline
\multirow{4}{*}{Gowalla}   & \multirow{2}{*}{BPRMF}       & R@20  &  \textbf{12.19}&  12.04\\
                            &                           & N@20 &  \textbf{7.76}&  7.62\\
                            &                           & Gini@20  &  98.02&  \textbf{96.31}\\ \cline{2-5} 
                            & \multirow{2}{*}{LightGCN} & R@20  &  \textbf{17.73}&  17.47\\
                            &                           & N@20 &  \textbf{11.16}&  11.01\\
                            &                           & Gini@20  &  98.17&   \textbf{96.48}\\ \hline
\end{tabular}
\end{table}

\subsubsection{Comparison of Fairness}
The performance comparison of item fairness is displayed in Table \ref{tab:ItemFairness}. We compare DVR-Gini with the BPR model based on BPRMF and LightGCN backbones. From the experimental findings, we draw the following conclusions: (i) DVR-Gini consistently outperforms the BPR model in terms of the Gini metric, as DVR-Gini utilizes Gini as the reward in its reinforcement learning-based optimization approach. By leveraging policy gradients learned from the Gini metric, the data valuator improves the fairness performance. (ii) The Recall@20 and NDCG@20 of DVR-Gini exhibit strong performance compared to the BPR model, showcasing the robustness of our reinforcement learning framework. This resilience is similar to the rationale discussed in the preceding section on diversity.

\begin{figure}[h]
    \centering
    \includegraphics[width=0.4\textwidth]{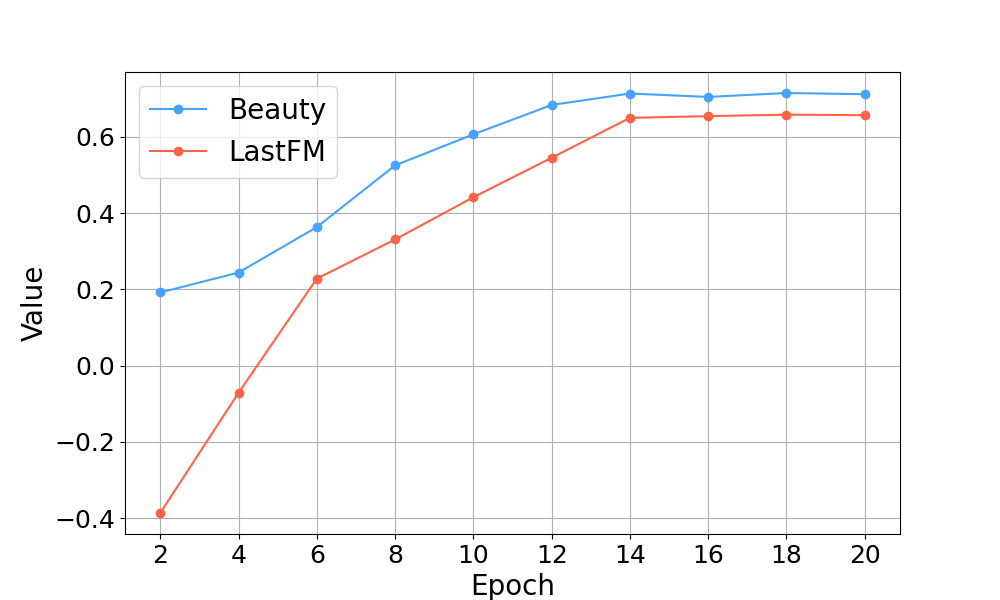}
    \caption{The Cosine Similarity between Negative BPR Loss and Shapley Value.}
    \label{fig:case_study}
\end{figure}

\subsubsection{Case Study of Shapley value}
To verify the interpretability of the assigned Shapley value, we conduct a case study of the BPR-Loss model on the Beauty and LastFM datasets. Figure \ref{fig:case_study} shows the average cosine similarity between the Shapley value and the negative BPR loss of the batch data in every two epochs. Our analysis revealed a progressive increase in cosine similarity throughout the training process. Notably, the Beauty and LastFM datasets exhibited distinct initial values, peak values, and growth rates. These variations could be attributed to differences in dataset size and density, resulting in varying convergence speeds.

%% file: Contents/7_conclusion.tex
\section{Conclusion}
We introduce an interpretable and adaptable recommendation framework to enhance data utilization and meet the specific requirements of model architectures and evaluation metrics. Our approach includes a data valuator to assess data quality through the calculation of Shapley value, ensuring robust mathematical properties. 
Additionally, we develop a metric adapter based on reinforcement learning to accommodate various evaluation metrics, including both differentiable and non-differentiable. Our extensive experiments on real-world datasets show that our framework outperforms state-of-the-art methods, achieving notable improvements in accuracy, diversity, and fairness metrics.

\section*{Acknowledgments}
This paper was supported by the NSF of Guangdong Province (Project No. 2024A1515010192), the Innovation and Technology Commission of Hong Kong (Project No. MHP/072/23). This paper was supported by the NSFC Young Scientists Fund (Project No. 9240127).

%% file: ltexpprt_doublecolumn.bbl
\begin{thebibliography}{10}

\bibitem{amari1993backpropagation}
{\sc S.-i. Amari}, {\em Backpropagation and stochastic gradient descent method}, Neurocomputing, 5 (1993), pp.~185--196.

\bibitem{caselles2018word2vec}
{\sc H.~Caselles-Dupr{\'e}, F.~Lesaint, and J.~Royo-Letelier}, {\em Word2vec applied to recommendation: Hyperparameters matter}, in Proceedings of the International Conference on Recommender Systems, 2018, pp.~352--356.

\bibitem{Celma:Springer2010}
{\sc O.~Celma}, {\em {Music Recommendation and Discovery in the Long Tail}}, Springer, 2010.

\bibitem{chen2023harsanyinet}
{\sc L.~Chen, S.~Lou, K.~Zhang, J.~Huang, and Q.~Zhang}, {\em Harsanyinet: computing accurate shapley values in a single forward propagation}, in Proceedings of the 40th International Conference on Machine Learning, 2023, pp.~4804--4825.

\bibitem{10.1145/3097983.3098202}
{\sc T.~Chen, Y.~Sun, Y.~Shi, and L.~Hong}, {\em On sampling strategies for neural network-based collaborative filtering}, in Proceedings of the International Conference on Knowledge Discovery and Data Mining, 2017, pp.~767--776.

\bibitem{cho2011friendship}
{\sc E.~Cho, S.~A. Myers, and J.~Leskovec}, {\em Friendship and mobility: user movement in location-based social networks}, in Proceedings of the International Conference on Knowledge Discovery and Data Mining, 2011, pp.~1082--1090.

\bibitem{10.5555/3367243.3367349}
{\sc J.~Ding, Y.~Quan, X.~He, Y.~Li, and D.~Jin}, {\em Reinforced negative sampling for recommendation with exposure data.}, in Proceedings of the International Joint Conference on Artificial Intelligence, 2019, pp.~2230--2236.

\bibitem{gantner2012personalized}
{\sc Z.~Gantner, L.~Drumond, C.~Freudenthaler, and L.~Schmidt-Thieme}, {\em Personalized ranking for non-uniformly sampled items}, in Proceedings of the International Conference on KDD Cup, 2011, pp.~231--247.

\bibitem{ghorbani2019data}
{\sc A.~Ghorbani and J.~Zou}, {\em Data shapley: Equitable valuation of data for machine learning}, in International Conference on Machine Learning, 2019, pp.~2242--2251.

\bibitem{harsanyi1982simplified}
{\sc J.~C. Harsanyi and J.~C. Harsanyi}, {\em A simplified bargaining model for the n-person cooperative game}, Papers in game theory,  (1982), pp.~44--70.

\bibitem{he2016ups}
{\sc R.~He and J.~McAuley}, {\em Ups and downs: Modeling the visual evolution of fashion trends with one-class collaborative filtering}, in Proceedings of the International Conference on World Wide Web, 2016, pp.~507--517.

\bibitem{he2020lightgcn}
{\sc X.~He, K.~Deng, X.~Wang, Y.~Li, Y.~Zhang, and M.~Wang}, {\em Lightgcn: Simplifying and powering graph convolution network for recommendation}, in Proceedings of the International Conference on Research and Development in Information Retrieval, 2020, pp.~639--648.

\bibitem{he2017neural}
{\sc X.~He, L.~Liao, H.~Zhang, L.~Nie, X.~Hu, and T.-S. Chua}, {\em Neural collaborative filtering}, in Proceedings of the International Conference on World Wide Web, 2017, pp.~173--182.

\bibitem{HidasiKBT15}
{\sc B.~Hidasi, A.~Karatzoglou, L.~Baltrunas, and D.~Tikk}, {\em Session-based recommendations with recurrent neural networks}, in Proceedings of the 4th International Conference on Learning Representations, 2015.

\bibitem{jarvelin2002cumulated}
{\sc K.~J{\"a}rvelin and J.~Kek{\"a}l{\"a}inen}, {\em Cumulated gain-based evaluation of ir techniques}, Transactions on Information Systems, 20 (2002), pp.~422--446.

\bibitem{koren2009matrix}
{\sc Y.~Koren, R.~Bell, and C.~Volinsky}, {\em Matrix factorization techniques for recommender systems}, Computer, 42 (2009), pp.~30--37.

\bibitem{li2020sampling}
{\sc D.~Li, R.~Jin, J.~Gao, and Z.~Liu}, {\em On sampling top-k recommendation evaluation}, in Proceedings of the International Conference on Knowledge Discovery \& Data Mining, 2020, pp.~2114--2124.

\bibitem{10.1145/3366423.3380187}
{\sc D.~Lian, Q.~Liu, and E.~Chen}, {\em Personalized ranking with importance sampling}, in Proceedings of The Web Conference, 2020, pp.~1093--1103.

\bibitem{ma2019hierarchical}
{\sc C.~Ma, P.~Kang, and X.~Liu}, {\em Hierarchical gating networks for sequential recommendation}, in Proceedings of the International Conference on Knowledge Discovery \& Data Mining, 2019, pp.~825--833.

\bibitem{ma2020memory}
{\sc C.~Ma, L.~Ma, Y.~Zhang, J.~Sun, X.~Liu, and M.~Coates}, {\em Memory augmented graph neural networks for sequential recommendation}, in Proceedings of the Conference on Artificial Intelligence, vol.~34, 2020, pp.~5045--5052.

\bibitem{10.1145/2556195.2556248}
{\sc S.~Rendle and C.~Freudenthaler}, {\em Improving pairwise learning for item recommendation from implicit feedback}, in Proceedings of the International Conference on Web Search and Data Mining, 2014, pp.~273--282.

\bibitem{10.5555/1795114.1795167}
{\sc S.~Rendle, C.~Freudenthaler, Z.~Gantner, and L.~Schmidt-Thieme}, {\em Bpr: Bayesian personalized ranking from implicit feedback}, in Proceedings of the International Conference on Uncertainty in Artificial Intelligence, 2009, pp.~452--461.

\bibitem{10.1145/3357384.3357895}
{\sc F.~Sun, J.~Liu, J.~Wu, C.~Pei, X.~Lin, W.~Ou, and P.~Jiang}, {\em Bert4rec: Sequential recommendation with bidirectional encoder representations from transformer}, in Proceedings of the 28th ACM International Conference on Information and Knowledge Management, 2019, p.~1441–1450.

\bibitem{8970709}
{\sc J.~Sun, Y.~Zhang, C.~Ma, M.~Coates, H.~Guo, R.~Tang, and X.~He}, {\em { Multi-graph Convolution Collaborative Filtering }}, in 2019 IEEE International Conference on Data Mining (ICDM), 2019, pp.~1306--1311.

\bibitem{10.1145/3437963.3441800}
{\sc W.~Wang, F.~Feng, X.~He, L.~Nie, and T.-S. Chua}, {\em Denoising implicit feedback for recommendation}, in Proceedings of the International Conference on Web Search and Data Mining, 2021, pp.~373--381.

\bibitem{10.1145/3366423.3380098}
{\sc X.~Wang, Y.~Xu, X.~He, Y.~Cao, M.~Wang, and T.-S. Chua}, {\em Reinforced negative sampling over knowledge graph for recommendation}, in Proceedings of the Web Conference, 2020, pp.~99--109.

\bibitem{winter2002shapley}
{\sc E.~Winter}, {\em The shapley value}, Handbook of game theory with economic applications, 3 (2002), pp.~2025--2054.

\bibitem{wu2022adapting}
{\sc H.~Wu, C.~Ma, Y.~Zhang, X.~Liu, R.~Tang, and M.~Coates}, {\em Adapting triplet importance of implicit feedback for personalized recommendation}, in Proceedings of the International Conference on Information \& Knowledge Management, 2022, pp.~2148--2157.

\bibitem{yoon2020data}
{\sc J.~Yoon, S.~Arik, and T.~Pfister}, {\em Data valuation using reinforcement learning}, in International Conference on Machine Learning, PMLR, 2020, pp.~10842--10851.

\bibitem{10.1145/3626772.3657748}
{\sc X.~Zhang, B.~Xu, Z.~Ren, X.~Wang, H.~Lin, and F.~Ma}, {\em Disentangling id and modality effects for session-based recommendation}, in Proceedings of the 47th International ACM SIGIR Conference on Research and Development in Information Retrieval, 2024, p.~1883–1892.

\bibitem{10.1145/3626772.3657761}
{\sc X.~Zhang, B.~Xu, Y.~Wu, Y.~Zhong, H.~Lin, and F.~Ma}, {\em Finerec: Exploring fine-grained sequential recommendation}, in Proceedings of the 47th International ACM SIGIR Conference on Research and Development in Information Retrieval, 2024, p.~1599–1608.

\bibitem{10.1145/3477495.3532043}
{\sc X.~Zhang, B.~Xu, L.~Yang, C.~Li, F.~Ma, H.~Liu, and H.~Lin}, {\em Price does matter! modeling price and interest preferences in session-based recommendation}, in Proceedings of the 45th International ACM SIGIR Conference on Research and Development in Information Retrieval, 2022, p.~1684–1693.

\bibitem{zhu2022gain}
{\sc Q.~Zhu, H.~Zhang, Q.~He, and Z.~Dou}, {\em A gain-tuning dynamic negative sampler for recommendation}, in Proceedings of the Web Conference, 2022, pp.~277--285.

\end{thebibliography}
